\documentclass{article}

\usepackage[nonatbib,preprint]{nips_2018}

\usepackage[utf8]{inputenc} 
\usepackage[T1]{fontenc}    
\usepackage{hyperref}       
\usepackage{url}            
\usepackage{booktabs}       
\usepackage{amsfonts}       
\usepackage{nicefrac}       
\usepackage{microtype}      

\usepackage{graphicx}
\usepackage{subfig}

\usepackage{algorithm}
\usepackage{algorithmic}

\usepackage{amssymb}
\usepackage{amsmath}
\usepackage{bm}
\newtheorem{defin}{Definition}
\newtheorem{theo}{Theorem}

\title{Solving Non-identifiable Latent Feature Models}

\author{
  Ryota Suzuki%
  \thanks{r-suzuki@fa.jp.nec.com}
  \AND
  Shingo Takahashi%
  \thanks{s-takahashi@fh.jp.nec.com}
  \AND
  Murtuza Petladwala%
  \thanks{murtuza@cq.jp.nec.com}
  \AND
  Shigeru Kohmoto%
  \thanks{s-koumoto@bq.jp.nec.com}
  \\
  \\
  NEC Data Science Research Laboratories \\
  Tokyo, Japan
}

\begin{document}

\maketitle

\begin{abstract}
  Latent feature models (LFM)s are widely employed for extracting latent structures of data.
  While offering high, parameter estimation is difficult with LFMs because
  of the combinational nature of latent features,
  and non-identifiability is a particularly difficult problem
  when parameter estimation is not unique
  and there exists {\it equivalent} solutions.
  In this paper, a necessary and sufficient condition for non-identifiability
  is shown. The condition is significantly related to dependency of features,
  and this implies that non-identifiability may often occur in real-world applications.
  A novel method for parameter estimation
  that solves the non-identifiability problem is also proposed.
  This method can be combined as a post-process with existing methods
  and can find an appropriate solution by {\it hopping} efficiently
  through equivalent solutions.
  We have evaluated the effectiveness of the method
  on both synthetic and real-world datasets.
\end{abstract}

\section{Introduction}
\label{sec:intro}

Latent variable models are widely used for obtaining hidden data-structures.
A mixture model is a leading example of such a latent variable model,
in which each instance of data is classified into a {\it latent class}.
Latent feature model (LFM) \cite{Griffiths2005} is an extension of a mixture model,
in which data is characterized into not only one
but into a combination of {\it latent features}.
LFM is used for various applications to extract such hidden structures of data
in medical \cite{Ruiz2014, Valera2017}, facial images \cite{Broderick2013}, 
and social, gene, and document networks \cite{Zhu2017}.


While LFM offers wide application,
parameter estimation (unsupervised learning) is difficult with it
because the optimization algorithms encounters its combinational non-convex nature;
possible combinations of $K$ features arising in $N$ data increase exponentially in $2^{NK}$,
and multi-modal structure of cost-functions
(such as log-likelihood and model evidence)
results in numerous local optima,
preventing optimization algorithms to obtain global optima.
Several approaches have been proposed to deal with non-convexity.
Reed and Zoubin \cite{Reed2013} focused on the submodularity, a discrete analog of convexity,
of a cost function in a non-negative assumption of features,
in which efficient greedy algorithm is available.
Another approach has been proposed by Yen et al. \cite{Yen2017},
in which a convex relaxation is employed with a {\it Lasso} regularizer,
and it may be solved as a certain class of semi-definite programming.
Even though these methods avoid the problem of non-convexity,
they still have to face the other difficulty, which is non-identifiability.

Identifiability in latent variable models represents the uniqueness of parameter estimation \cite{Murphy2012}.
The parameters are correctly estimated if the solution of an optimization problem is unique.
However, if there are multiple parameters that result in the same cost,
the solution of the optimization problem may not be unique.
In such a {\it non}-identifiable situation,
error in parameter estimation may significantly worsen
\cite{Watanabe2001}.
A special case of non-identifiability in LFM is shown in \cite{Hayashi2013},
in which two features have the same value,
and their method solves the problem by balancing the size of the features.
Additionally, \cite{Yen2017} have shown a sufficient condition for identifiability
and also that the condition holds with high probability 
under an assumption that
features appear in an {\it independently and identically distributed} (i.i.d.)
{\it Bernoulli} process.
However, i.i.d. {\it Bernoulli},
assumed in most existing works \cite{Hayashi2013, Tung2014}
including an {\it Indian buffet process} (IBP)
\cite{ Griffiths2005, Broderick2013, Reed2013, Doshi-velez2009, Griffiths2011},
is too stringent an assumption in many real-world applications;
there might be hidden constraints for which some features do not appear at the same time,
and/or features may have a hidden hierarchical structure in which one subsumes another.
We show that, in such a case of features' having a dependency,
parameter estimation is non-identifiable,
and optimization methods may face the difficulty.
Another case of non-identifiability we show is that of the existence of {\it bias},
which can be represented as a feature commonly appearing in all the data
\cite{Ruiz2014,Valera2017}.

Since difficulty in non-identifiability is due to the existence of solutions having the same cost,
which we refer to here as {\it equivalent solutions},
optimization methods may find a solution that is not the true parameter but {\it equivalent} one.
We have developed a {\it hopping} algorithm that efficiently finds equivalent solutions
from one to another in succession,
which maximizes prior probabilities without degrading likelihood.


In this paper, we present a solution to the problem of non-identifiability,
and our contributions are as follows:
first, we have derived a necessary and sufficient condition for non-identifiability in LFM.
Secondly, we have derived sufficient conditions for non-identifiability that is,
significantly, related to the dependency of features.
Thirdly, we have developed a novel method for parameter estimation
that can be combined as a post-process with existing methods
and that can find appropriate solutions by {\it hopping} through {\it equivalent} solutions.
Finally, we have also shown the effectiveness of the new method
on both synthetic and real-world datasets.

\section{Latent Feature Models}
\label{sec:model}
In a latent feature model (LFM), observed data is assumed to be
represented as a combination of $K$ latent features.
Let $X \in \mathbb{R}^{N\times D}$ be an observed data matrix,
and its rows represent instances of $D$-dimensional observations.
We assume that $X$ can be generated as:
\begin{equation}
  \label{eq:mf-lfm}
  X = ZW + \varepsilon, \;\;\; Z \sim P_Z(Z),\;\;\; W \sim P_W(W)
\end{equation}
where $k$-th row of $W \in \mathbb{R}^{K \times D}$ is the $k$-th latent feature $\bm{w}_k$,
and the $k$-th column $\bm{z}_k$ of unknown binary matrix $Z \in \{0,1\}^{N\times K}$
represents the incidence of the $k$-th latent feature along $N$ data,
and $\varepsilon$ is an unknown noise matrix.
A typical class of LFMs is linear-Gaussian LFMs \cite{Hayashi2013, Tung2014}:
\begin{equation}
  \label{eq:typical-lfm}
  P_Z = \mathrm{Bernoulli}(\bm{\pi}), \;
  P_W = \mathcal{N}(0, \sigma_W^2), \;
  \varepsilon \sim \mathcal{N} (0, \sigma_X^2).
\end{equation}
where $\bm{\pi}=(\pi_1,\cdots,\pi_K)$ is probability for feature incidence.
If $\pi_k=1/2$, a maximum a posteriori (MAP) inference is obtained from following optimization problem:
\begin{equation}
  \label{eq:typical-map}
  \mathop{\mathrm{arg~min}}_{Z \in\{0,1\}^{N\times K}, W\in \mathbb{R}^{K\times D}}
  \left\{ \| X-ZW \|_F^2 + \tau \| W \|_F^2 \right\},
\end{equation}
where $\tau = \sigma_X^2 / \sigma_W^2$,
and it becomes maximum likelihood (ML) estimation in $\tau \to 0$.
IBP is also used for $P_Z$ in Bayesian nonparametric settings
\cite{Griffiths2005, Broderick2013, Reed2013, Doshi-velez2009, Griffiths2011},
and the optimization problem (\ref{eq:typical-map}) can also be obtained
from MAP asymptotics of IBP linear-Gaussian LFM with $\ell_2$ regularization
\cite{Broderick2013, Yen2017}.

From a general point of view, LFM (\ref{eq:mf-lfm})
can be seen as a class of matrix factorizations of $X$ into $Z$ and $W$
with a constraint such that $Z$ is binary \cite{Slawski2013}.
Our analyses and methods deal not only with typical LFMs
but also with such a general class of matrix factorization.
Factorial hidden Markov Models (FHMMs) \cite{Ghahramani1997,Gael2008}
and non-negative LFMs \cite{Reed2013,Omalley2017}
are an example of such a class of matrix factorization,
where additional constraints and prior probabilities are assumed.
We focus on a general characteristics of matrix factorization (\ref{eq:mf-lfm}),
and our analyses and methods are applicable to a wide-range of matrix factorization problems.

Most existing works on LFMs,
explicitly or implicitly,
make two strong assumptions related to identifiability.
The first common but strong assumption is the statistical independence of
features \cite{Broderick2013, Reed2013, Hayashi2013, Tung2014, Doshi-velez2009, Griffiths2011}.
In these case, one assumes that the incidence of an individual feature
will be independently generated in a Bernoulli process.
As we will see in Sec. \ref{subsec:sufficient-conditions},
the independence of features is sometimes too strong an assumption in actual situations,
and an absence of independence will cause the problem of non-identifiability.

The second assumption related to identifiability is that
a model has zero bias $\mathbb{E}[\varepsilon|Z] = 0$
\cite{Broderick2013, Reed2013, Hayashi2013, Tung2014, Doshi-velez2009, Griffiths2011, Slawski2013}.
There might, however, be a background feature
common to all instances in actual situations.
Introducing bias term $\bm{w}_\mathrm{bias}$ to (\ref{eq:mf-lfm}) is equivalent to
an additional feature that is always active \cite{Ruiz2014},
and the existence of such a bias results in non-identifiability,
as we will see in Sec. \ref{subsec:sufficient-conditions}.

\section{Analyses of Non-identifiability}
\label{sec:identifiability}

\subsection{Non-identifiability in LFMs}

\begin{figure}[tb]
  \begin{center}
    \centerline{ \includegraphics[width=\textwidth]{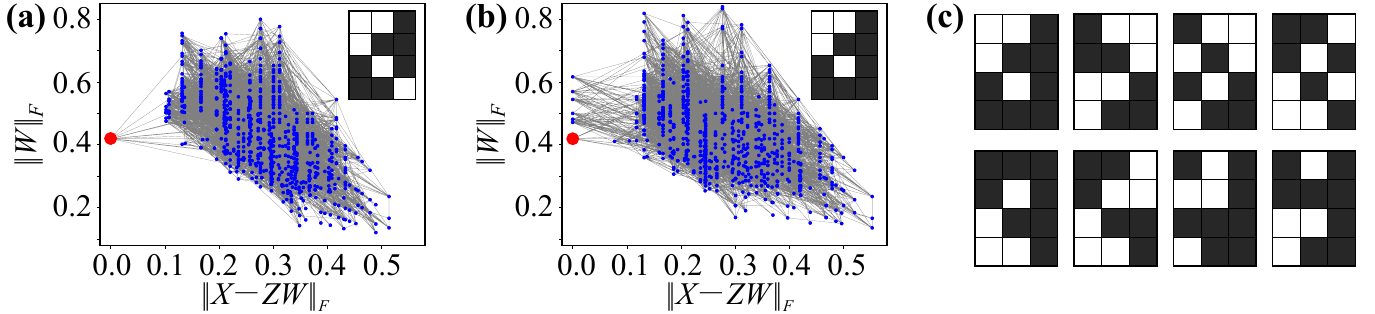} }
    \vspace{-5pt}
    \caption{
      Whole feasible solutions of $Z\in \{0,1\}^{4\times 3}$ for 
      {(\bf a)} identifiable and {(\bf b)} non-identifiable cases.
      Blue dots correspond to individual solutions
      that are connected to one another by gray lines of Hamming distance one.
      Ground truth $Z^*$ is plotted as a red circle and its components are shown in insets.
      {\bf (c)} shows solutions equivalent to the ground truth of {\bf (b)}.
      We use noiseless data with features $W^*$, as three of four synthetic images
      originally used in \cite{Griffiths2005}.
      \label{fig:feasible_space}
    }
    \vspace{-20pt}
  \end{center}
\end{figure}

In this paper,
as is also seen in \cite{Slawski2013, Yen2017},
we consider identifiability in LFMs in terms of uniqueness of the solution.
Before presenting our theoretical analyses,
let us briefly overview (non-)identifiability in LFMs with an example.

In LFMs, non-identifiability difficulties are mainly due to discrete nature of $Z$.
Since the optimization problem (\ref{eq:typical-map}) for $W$ with fixed $Z$ is convex,
the MAP solution $\hat{W}$ corresponding to $\hat{Z}$
(which is also an expectation $\mathbb{E}[W|\hat{Z},X]$)
is uniquely calculated in closed form \cite{Griffiths2005, Yen2017}.
The uniqueness of $Z$ is more complicated.
Figure~\ref{fig:feasible_space} (a, b) shows every possible combinations of
$Z$ in two different $X$'s.
The axes of the figures correspond to the terms of (\ref{eq:typical-map}).
In the identifiable case (a),
only the ground truth has zero-residual $\|X-ZW\|_F = 0$.
However, in the non-identifiable case (b),
there are many solutions having zero-residual,
and inference methods minimizing the residual, 
such as \cite{Yen2017, Tung2014},
may fall into incorrect zero-residual solution rather than ground truth.
Moreover, zero-residual solutions (Fig. \ref{fig:feasible_space}c)
are distant from one another at Hamming distances,
and this results in a multi-modal nature in a cost function preventing iterative algorithms,
such as Markov-chain Monte Carlo (MCMC)\cite{Griffiths2005,Doshi-velez2009}
and variational Bayesian (VB)\cite{Doshi-velez-variational2009} methods,
from converging to a global optimum.


\subsection{General Matrix Factorizations}

Let us next consider conditions for (non-)identifiability
and properties of {\it equivalent} solutions.
Since LFMs are a class of matrix factorizations,
we start from identifiability in general matrix factorization
\cite{Laurberg2008, Yen2017, Slawski2013} 
to get an overall picture.
\begin{defin}[Identifiability]
  \label{defin:ident}
  \vspace{-3pt}
  Let $\mathcal{Z} \subset \mathbb{R}^{N\times K}, \mathcal{W} \subset \mathbb{R}^{K\times D}$ be sets of matrices.
  We say that a pair of matrices $(Z, W) \in \mathcal{Z}\times \mathcal{W}$ is identifiable
  if for all $(Z', W') \in \mathcal{Z} \times \mathcal{W}$,
  $ZW = Z'W'$ implies
     $\{Z_{:,k}\}_{k=1}^K = \{Z'_{:,k}\}_{k=1}^K$ and
     $\{W_{k,:}\}_{k=1}^K = \{W'_{k,:}\}_{k=1}^K$.
  \vspace{-3pt}
\end{defin}
We can consider wide-ranging classes of matrix factorization
by choosing $\mathcal{Z}$ and $\mathcal{W}$.
The LFM is the case of $\mathcal{Z} = \{0,1\}^{N \times K}$ and $\mathcal{W} = \mathbb{R}^{K \times D}$.
We can further say that a matrix factorization of $X \in \mathbb{R}^{N\times D}$ is
identifiable if one of the minimizers of a residual $\| X-ZW \|_F^2$ (i.e., one of the ML solutions)
is identifiable.
In noiseless settings ($\varepsilon\!=\!0$), the identifiability of $X$ is consistent with
that of the ground truth $(Z^*,W^*)$.

It is known that most equivalent solutions of matrix factorization $Z'W'=ZW$ have a specific form.
For a context of non-negative matrix factorization (NMF),
where $\mathcal{Z} = \mathbb{R}_+^{N \times K}$ and $ \mathcal{W} = \mathbb{R}_+^{K \times D}$,
Laurberg, et al. \cite{Laurberg2008} showed that all equivalent NMF solutions $Z'W'=ZW$ have a form
$Z'=ZU, W'=U^{-1}W$ if $\mathrm{rank}(ZW) = K$.
This assertion is also true in the general matrix factorization in Definition~\ref{defin:ident}
since their proof only used properties of linear space.
\begin{theo} {\bf Laurberg, et al. \cite{Laurberg2008}}
  \label{theo:laurberg}
  Let $Z \in \mathcal{Z} \subset \mathbb{R}^{N\times K}$ and
  $W \in \mathcal{W} \subset \mathbb{R}^{K\times D}$.
  Assume $\mathrm{rank}(ZW) = K$.
  Then for any $(Z', W')\in \mathcal{Z} \times \mathcal{W}$, $ZW = Z'W'$ holds if and only if
  there exists a regular matrix $U \in \mathbb{R}^{K \times K}$
  such that $Z'=ZU, W'=U^{-1}W$.
\end{theo}

\subsection{Equivalence Classes}

Let us consider a set of equivalent solutions of an LFM to quantify (non-)identifiability.
Let $[(Z,W)]$ be an equivalence class of an equivalence relation "$\sim$" defined as $(Z,W)\sim(Z',W')\,\Leftrightarrow\,ZW=Z'W'$.
Namely,
$ [(Z,W)] = \left\{ (Z', W') \in \mathcal{Z}\times\mathcal{W} \middle| ZW = Z'W' \right\}$.

Assuming $\mathrm{rank}(ZW) = K$,
Theorem~\ref{theo:laurberg} guarantees that all elements of $[(Z,W)]$
will be represented as $({Z}U, U^{-1}{W})$.
Then we can consider LFM
($\mathcal{Z}\!=\!\{0,1\}^{N \times K}, \mathcal{W}\!=\! \mathbb{R}^{K\times D}$),
the equivalence class to be expressible as
$ [(Z, W)] = \left\{ ({Z}U, U^{-1}{W} ) \middle| U \in H({Z}) \right\} $,
where
\vspace{ -3pt}
\begin{equation*}
  H({Z}) = \left\{ U \in \mathbb{R}^{K \times K} \middle| \det U \neq 0, {Z}U \in \{0,1\}^{N \times K} \right\}.
\end{equation*}
\vskip -5pt
Non-identifiability in the LFM is characterized by
distinct elements $U\in H(Z)$ other than permutation matrices
$U \not \in \mathcal{S}_K$, where $\mathcal{S}_K$ is symmetric group of degree $K$.
To reduce the degree of freedom in permutation,
we assume a quotient set\footnote{
  In a precise sense, $H(Z) / \mathcal{S}_K$ stand s for
  $H(Z) / \mathop{\sim_{\mathcal{S}_K}} $,
  the quotient of $H(Z)$ by an equivalence relation ``$\sim_{\mathcal{S}_K}$'' such that
  $U \mathop{\sim}_{\mathcal{S}_K} U'
  \Leftrightarrow {}^\exists \sigma \in\mathcal{S}_K, U'=U\sigma$.
} $H(Z) / \mathcal{S}_K$,
which is a set of sets, in which each set consists of transform matrices $U \in H(Z)$
having the same column entries but in different orders.

\subsection{Equivalent Condition for Identifiability in LFMs}

By using the notation above,
non-identifiability can be quantified by
a cardinality $|H(Z)/\mathcal{S}_K|$,
which represents the distinct number of equivalent solutions
avoiding duplication in permutations.
Additionally, we get
$|H(Z)/\mathcal{S}_K|=1$ as
a necessary and sufficient condition
for the identifiability under the assumption of $\mathrm{rank}(ZW)\!=\!K$.
We consider $H(Z)/\mathcal{S}_K$ to be {\it trivial} if $H(Z)/\mathcal{S}_K$
has only one element
.
While the assumption $\mathrm{rank}(ZW)\!=\!K$ denotes both
$\mathrm{rank}(Z)\!=\!K$ and $\mathrm{rank}(W)\!=\!K$,
we further prove that the former is unnecessary,
and we get a following theorem providing a 
necessary and sufficient condition for the identifiability:
\begin{theo}
  \label{theo:ident}
  Let $Z \in \{0,1\}^{N\times K}, W \in  \mathbb{R}^{K\times D}$.
  Assume $\mathrm{rank}(W) = K$.
  Then $(Z, W)$ is identifiable if and only if $H({Z}) / \mathcal{S}_K$ is trivial.
\end{theo}
Note that Theorem~\ref{theo:ident} is a stronger result
than the ``{\it identifiability condition}'' mentioned in \cite{Yen2017},
which supplies only a sufficient condition for identifiability
in the case of $\mathrm{rank}(ZW)\!=\!K$.
While the assumption $\mathrm{rank}(W)\!=\!K$ in Theorem~\ref{theo:ident}
is not necessarily required for the
identifiability\footnote{
  We can show an example in which identifiability holds despite $\mathrm{rank}(W)\!<\!K$. See Supplemental Materials.
},
we assume $\mathrm{rank}(W) = K$ in the following discussion
in consideration for $D \gg K$ in many applications.

\begin{figure}[tb]
\begin{center}
    \centerline{\includegraphics[width=\textwidth]{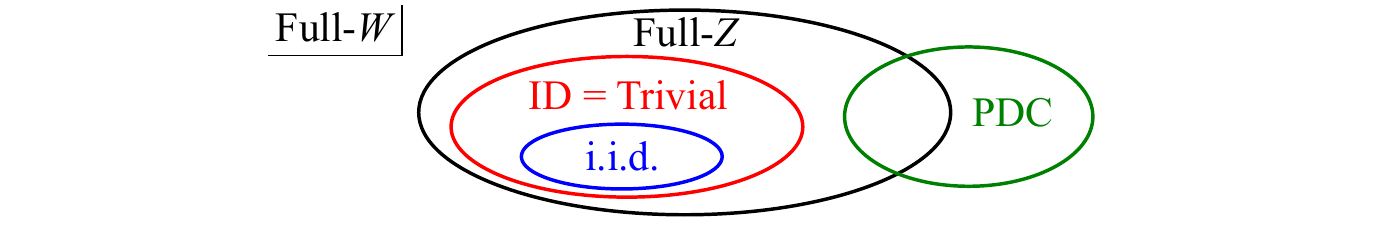}}
    \caption{
      Relationship between conditions under the assumption $\mathrm{rank}(W) = K$ ({\bf Full-W}).
      {\bf Full-Z}: $\mathrm{rank}(Z)=K$,
      {\bf ID}: Identifiability in LFMs (Definition \ref{defin:ident}),
      {\bf Trivial}: $|H(Z)/\mathcal{S}_K| = 1$ (Theorem~\ref{theo:ident}),
      {\bf i.i.d.}: $Z \sim \mathrm{i.i.d.}\,Bernoulli(p)$
      with $0<p<1, N \to \infty$ (Theorem~\ref{theo:iid}),
      {\bf PDC}: pairwise dependency conditions (Theorem~\ref{theo:pdc}).
      \label{fig:conditions}
    }
    \vspace{-20pt}
  \end{center}
\end{figure}

Finally, in Figure~\ref{fig:conditions},
we summarize the relationship between the conditions
we have derived:
({\bf Full-Z}) $\mathrm{rank}(Z)=K$,
({\bf ID})~the identifiability,
and ({\bf Trivial}) $|H(Z)/\mathcal{S}_K|=1$.

\subsection{Sufficient Conditions for (Non-)Identifiability}
\label{subsec:sufficient-conditions}
\vspace{-2pt}


Non-identifiability in LFMs is significantly related to dependency between features.
We illustrate here two sufficient conditions,
both for identifiability and non-identifiability
({\bf i.i.d.} and {\bf PDC} in Figure~\ref{fig:conditions}),
related to independence and dependence of features, respectively.

The first condition, the sufficient condition for identifiability,
is the statistical independence of the features.
\cite{Yen2017} have shown that an LFM is identifiable with high probability
if entries of $Z$ are i.i.d. {\it Bernoulli} $p\!=\!0.5$.
More generally, identifiability holds with probability one
in $N\!\to\!\infty$ for any $0\!<\!p\!<\!1$.
We show this in the following theorem:
\begin{theo}
  \label{theo:iid}
  $H({Z}) / \mathcal{S}_K$ is trivial if 
  $Z$ is a binary matrix s.t. $\{ Z_{n,:} \}_{n=1}^{N} = \{0,1\}^{K}$.
\end{theo}
If $Z$ is i.i.d. {\it Bernoulli} with $0<p<1$,
probabilities for every combination of each row$P(Z_{n,:})$ will be non-zero.
Therefore, in the limit of $N \to \infty$,
all the combinations $\{0,1\}^K$ may appear in rows of $Z$ with probability one,
and this results in identifiability via Theorem~\ref{theo:iid}.
In other words, from a contraposition of Theorem~\ref{theo:iid},
it can be said that possible non-identifiability is due to a lack of observed combinations.
In real-world applications
observing all $2^K$ combinations is rarely possible,
and, as we will see later, some combinations may never appear in rows of $Z$ even if
$N \to \infty$ because of hidden dependency between features.

The second condition is in regard to non-identifiability.
We propose three {\it pairwise dependency conditions} (PDCs)
sufficient for non-identifiability,
\begin{theo}
  \label{theo:pdc}
  Let ${Z} \in {0,1}^{N \times K}$ be a binary matrix.
  $|H({Z})/\mathcal{S}_K| \geq 3$ holds if
  there exists a distinct pair $i,j \;(i\neq j)$ of features
  satisfying one of following conditions for all $n = 1,\cdots,N$:
  \begin{itemize}
    \vspace{-5pt}
    \setlength{\itemindent}{-10pt}
    \setlength{\itemsep}{-4pt}
    \item {\rm PDC1:} $(z_{n,i} = 1) \Rightarrow (z_{n,j}=0)$.
    \item {\rm PDC2:} $(z_{n,i} = 1) \Rightarrow (z_{n,j}=1)$.
    \item {\rm PDC3:} $(z_{n,i} = 0) \Rightarrow (z_{n,j}=0)$.
    \vspace{-3pt}
  \end{itemize}
\end{theo}
The PDCs in Theorem~\ref{theo:pdc} often appear in real-world applications\footnote{
  We can show the commonality of PDCs from our survey on actual datasets. See Supplemental Materials.
},
and, unfortunately, they are unknown in most cases.
Inference methods may then suffer from non-identifiability
whenever there exists at least a pair of features PDC holds.
For instance, a typical case of PDC1 would be disjoint features.
Assuming, for example, that these features correspond to characteristics of cats,
then a pair of features $(i,j)= ("\textrm{male}", "\textrm{black}")$
may appear at the same time, but $(i,j)= ("\textrm{male}", "\textrm{female}")$
will not appear concurrently.
In the case of PDC2 and 3, typical cases would be latent hierarchical structures.
For example, in considering features $(i,j)= ("\textrm{cats}", "\textrm{mammals}")$,
the feature "mammals" is always active whenever "cats" is active since cats are mammals.

Another example of non-identifiability is the existence of a bias term.
An LFM with bias term is equivalent to an unbiased LFM with an extra feature
that is always active \cite{Ruiz2014, Valera2017}.
The existence of such a {\it bias feature} is followed by PDC2
because the feature is always active regardless of other features.
Let $(Z', W')$ be an equivalent solution corresponding to
the transform matrix
$U = I + \bm{e}_{i} \bm{e}_\mathrm{bias}^\mathrm{T} - 2 \bm{e}_i \bm{e}_i^\mathrm{T}$,
where the index "bias" refer to the bias feature,
absence and the presence of $i$-th feature $\bm{z}_i'$ is inverted from $\bm{z}_i$
and a sign of the $i$-th feature is flipped $\bm{w}_i' = -\bm{w}_i$
in swap of raising the level of the bias
$\bm{w}_\mathrm{bias}' = \bm{w}_\mathrm{bias} + \bm{w}_i$.
Such an ``inverted'' solution has been obtained by some algorithms,
including Gibbs sampler \cite{Ghahramani2006},
K-features \cite{Broderick2013}, and possibly other LFM algorithms.
In the case of a biased LFM,
we can prove that there is a lower bound to the number of equivalent solutions
$|H(Z)/\mathcal{S}_K| \geq (K+1) \times 2^{K-1}$ (see Supplemental Materials).

\vspace{-3pt}
\section{Hopping through Equivalent Solutions}
\label{sec:method}
\vspace{-3pt}

We now consider finding a superior solution among equivalent solutions.
Once an arbitrary estimator finds a (not necessarily optimal) MAP solution $(\hat{Z}, \hat{W})$
for the model \eqref{eq:mf-lfm},
there might be some equivalent solutions $(Z',W') \in [(\hat{Z},\hat{W})]$.
Although they have the same residual,
some of those might be close to the ground truth $(Z^*, W^*)$ but others might be far from it.
Our method obtains a superior one among them having a maximal prior probability.
In it, for efficiency, we assume transform matrix $U$ to be an integer matrix,
i.e., we sample $U$ from a subset of $H(\hat{Z})$:
\vspace{ -5pt}
\begin{equation}
  \label{eq:Htilde}
  \tilde{H}(\hat{Z}) = H(\hat{Z}) \cap \mathbb{Z}^{K \times K}.
\end{equation}
\vskip -5pt
Although there might be $U \in H(Z)$ that is not an integer matrix in some cases,
we can show that $\tilde H(Z)=H(Z)$ holds in many cases in consideration of $N \gg K$ (see Supplemental Materials).

Let us next introduce a quadratic form
$f(Z)= \frac 12 \sum_{n,k}z_{n,k}(z_{n,k}-1)$,
then (\ref{eq:Htilde}) is denoted as
$\tilde{H}(Z) = \left\{ U \in \mathbb{Z}^{K \times K} \middle| \det U \neq 0, f(ZU)= 0 \right\}$.
We can assume $f(ZU)$ to be a cost function that measures
how different $ZU$ is from an binary matrix
since it is non-negative integer and becomes zero if and only if $ZU$ is a binary.
Further, $f(ZU)$ is an upper bound to the number of $ZU$ components other than 0 or 1,
and is equal to it if $-1 \preceq ZU \preceq 2$.

\subsection{Sampling Equivalent Solutions}

\begin{algorithm}[tb]
  \caption{Sampling Columns of~~$U$}
  \label{alg:candidate-sampler}
  \begin{algorithmic}
    \STATE {\bfseries Input:} binary matrix $Z$, sample size $N_s$
    \STATE Initialize $\mathcal U = \emptyset$.
    \STATE $\Phi, \Sigma, \Psi^\mathrm{T} = svd(Z)$;~~
           $\Lambda = \Psi \Sigma^{-1}$;~~
           $\bm{\mu} = \frac 12 \Lambda^\mathrm{T} Z^\mathrm{T} \bm{1}$.
    \FOR{$i=1$ {\bfseries to} $N_s$}
    \STATE Sample $f^{(i)}$ from arbitrary distribution.
    \STATE Sample $\bm{s}^{(i)}$ uniformly from a $(K\!-\!1)$-sphere of
      center $\bm{\mu}$ and radius $\sqrt{\| \bm{\mu} \|^2 + 2\{f^{(i)}\}^2 }$.
    \STATE $\bm{u}^{(i)} = \mathrm{round}(\Lambda \bm{s}^{(i)} )$
    \STATE $\mathcal{U} = \mathcal{U} \cup \{ \bm{u}^{(i)} \}$
      ~~{\bf if}
      ~~$ \bm{u}^{(i)} \neq \bm{0}$.
    \ENDFOR
  \end{algorithmic}
\end{algorithm}
\begin{algorithm}[tb]
  \caption{Equivalence Hopper}
  \label{alg:mcmc}
  \begin{algorithmic}
    \STATE {\bfseries Input:} initial matrices $Z$, $W$, \;
           $\mathcal{U}$ sampled by Algorithm \ref{alg:candidate-sampler}
    \STATE Initialize $U_{0} = I$
    \FOR{$m = 1,\cdots, M$}
    \STATE $J = \left\{ U' := U_{m-1} + (\bm{u}^{(i)} - \bm{u}_k) \bm{e}_k^\mathrm{T} \; \middle| \;
      \bm{u}^{(i)} \in \mathcal{U}; \;  k = 1,\!\cdots,\!K; \; \mathrm{rank}(U') = K
      \right\}$
    \STATE Sample  $U_{m} \in J$ according to the probability \eqref{eq:prob}.
    \ENDFOR
  \end{algorithmic}
\end{algorithm}
Let $\bm{u}_k$ be the $k$-th column of $U$,
and $\Lambda$ be a $K\times K$ matrix such that $ \Lambda\Lambda^\mathrm{T} = (Z^\mathrm{T} Z)^{-1} $,
which is calculated by {\it e.g.}, singular value decomposition (SVD) of $Z$,
$\bm{s}_k = \Lambda^{-1} \bm{u}_k$, and $\bm{\mu} = \frac 12 \Lambda^\mathrm{T} Z^\mathrm{T} \bm{1}$,
where $\bm{1} = (1,\cdots,1)^\mathrm{T}$.
then $f(ZU)$ for $k$-th column can be evaluated as:
\begin{equation}
  \label{eq:f}
  f(Z\bm{u}_k)
    = \frac 12 \left( \bm{u}_k^\mathrm{T} Z^\mathrm{T} Z \bm{u}_k 
        - \bm{u}_k^\mathrm{T} Z^\mathrm{T} \bm{1} \right)
    = \frac 12 \| \bm{s}_k - \bm{\mu} \|^2 - \frac 12 \| \bm{\mu} \|^2.
\end{equation}
If $U \in \tilde{H}(Z)$, (\ref{eq:f}) is evaluated to be zero,
then $\bm{s}_k$ will be on the ($K\!-\!1$)-sphere
of center $\bm{\mu}$ and radius $\|\bm{\mu}\|$.
A possible $\bm{u}_k$ can then be obtained by sampling
$\bm{s}$ uniformly from the ($K\!-\!1$)-sphere,
taking the nearest integer $\bm{u}\!=\!\mathrm{round}(\Lambda\bm{s})$,
and accept $\bm{u}_k = \bm{u}$ \,
if \, $ \bm{u}\neq \bm{0} \wedge f(Z\bm{u})\!=\!0 $.

The {\it strict} method mentioned above
(which samples $U$ strictly from $\tilde{H}({Z})$)
may, unfortunately, fail in some cases because ${Z}$, estimated with an arbitrary algorithm, may have randomness
and its flipped component may spoil equivalent solutions by breaking PDCs in Theorem~\ref{theo:pdc}.
To handle such a randomness, we employ a tolerance to the equivalent conditions.
If the transformed matrix $ Z' =  Z U$ includes an integer other than 0 or 1,
it will be rejected from equivalent solutions.
However, if there exists another binary matrix $\tilde Z'$ which is {\it close} to $ Z'$,
then $\tilde Z'$ will be a {\it nearly equivalent} solution to $ Z$.
We measure this closeness by $f(ZU)$ since it is an approximation for the number of
non-binary components as mentioned above,
and it can be calculated by (\ref{eq:f}) without scanning all $N$ rows of $ZU$.
In the 
Algorithm~\ref{alg:candidate-sampler} shows this {\it tolerant} method.
We first sample $f^*$ from some distribution,
and then sample $\bm{u}_k$ such that $f(Z\bm{u}_k) = f^*$
by sampling $\bm{s}$ uniformly from the ($K\!-\!1$)-sphere
of radius $\sqrt{\| \bm{\mu} \|^2 + {2f^*}^2 }$ instead of $\| \bm\mu \|$.
By choosing the distribution of $f^{*}$,
we can tune the tolerance for the protrusion of ${Z}U$ from binary matrices.
We employ discrete exponential distribution with parameter $\lambda$ for this role.
The {\it strict} case is a limit of $\lambda \to \infty$.

\begin{figure}[tb]
  \begin{center}
    \centerline{\includegraphics[width=\textwidth]{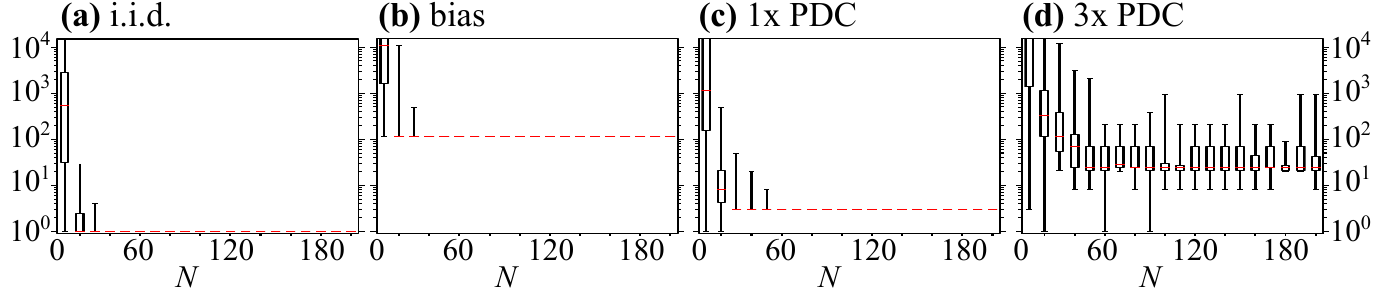}}
    \vspace{-3pt}
    \caption{
      Number of equivalent solutions $|\tilde H(Z) / \mathcal{S}_K|$ found by Algorithm~\ref{alg:candidate-sampler} for $K=6$.
      $Z$ is sampled from {\bf (a)} i.i.d {\it Bernoulli}(0.5),
      {\bf (b)} with a bias feature,
      and {\bf (c,d)} with 1, 3 pair(s) of features with PDC, respectively.
      Boxes, whiskers and red lines represent, respectively, quartiles, extrema, and medians
      of 50 trials.
      \label{fig:num_equiv}
    }
    \vspace{-25pt}
  \end{center}
\end{figure}

Once $\mathcal{U}$ is sampled by Algorithm \ref{alg:candidate-sampler}, equivalent solutions are
obtained by selecting $K$ columns of $U$ from $\mathcal{U}$ so that $U$ is a regular matrix.
Figure~\ref{fig:num_equiv} shows the number of equivalent solutions found from
$\tilde{H}(Z) / \mathcal{S}_K$ by using the {\it strict} version ($\lambda \to \infty$) of
Algorithm~\ref{alg:candidate-sampler}.
In the i.i.d. case (Fig.~\ref{fig:num_equiv}a),
the number of equivalent solutions diminishes rapidly to one with an increasing $N$.
This is consistent with results mentioned in \cite{Yen2017}.
Figure~\ref{fig:num_equiv} (b) to (d) show cases of
$Z$ sampled with a bias feature and PDCs (Theorem~\ref{theo:pdc}).
In these three cases, solutions remains multiple even in $N=200$,
keeping the problem non-identifiable.
In the case of a bias feature (Fig.~\ref{fig:num_equiv}b),
the number of solutions converges to the theoretical lower bound $(K+1)\times 2^{K-2} = 112$,
and when a pair of features holds a PDC (Fig.~\ref{fig:num_equiv}c), it converges to 3,
the lower bound shown in Theorem~\ref{theo:pdc}.
When three pairs of features have PDCs (Fig.~\ref{fig:num_equiv}d),
the number of solutions fluctuate up to $10^3$
while a combination of 3 independent PDCs will result in $3^3 = 27$ solutions.
This implies that multiple PDCs sharing a same feature give rise to another PDC,
e.g., PDC2 for $(i,j) = (1,2)$ and $(2,3)$ implies PDC2 for $(1,3)$.

\subsection{Optimizing over Equivalent Solutions}

Now we consider
obtaining appropriate solution among equivalent solutions.
We select $U \in \tilde{U}(\hat{Z})$ so that obtained solution
$(Z', W')=(\hat{Z}U, U^{-1}\hat{W})$ is more appropriate,
i.e., having higher prior probability without degrading likelihood.
So we employ the following cost function:
\begin{equation}
  \label{eq:cost}
  g(U; \hat{Z}, \hat{W}) = \log P_Z(\hat{Z} U) + \log P_W(U^{-1} \hat{W}) + \gamma f(\hat{Z}U).
\end{equation}
The first two terms of \eqref{eq:cost} correspond to log-priors in \eqref{eq:mf-lfm}.
The second term with parameter $\gamma$
plays a role keeping
$({Z}', {W}')$ nearly equivalent to $(\hat{Z}, \hat{W})$
and maintaining little change in the likelihood.

We can solve the optimization problem by MCMC-based sampling method (Algorithm~\ref{alg:mcmc}),
sampling $U$ by updating its columns successively according to a Boltzmann distribution with parameter $\beta$:
\begin{equation}
  \label{eq:prob}
  P(U|\hat{Z},\hat{W}) \propto \exp\{-\beta g(U;\hat{Z}, \hat{W})\},
\end{equation}
In the limit of $\beta \to \infty$, Algorithm \ref{alg:mcmc} becomes a greedy local-search algorithm,
which has a lower-bound in its optimality for typical linear-Gaussian LFMs (see Supplemental Materials).
However, sampling method ($\beta < \infty$) works well in many cases since the feasible space is small enough.

\section{Experiments}
\label{sec:experiments}


In this section, we demonstrate utility of our method by
applying it to both synthetic and actual data
as a post-process combined with existing algorithms.
Once an estimation $(\hat{Z},\hat{W})$ is obtained by such an algorithm,
we execute the Equivalence Hopper (Algorithm~\ref{alg:mcmc}) with $(\hat{Z}, \hat{W})$ as an input.
We employed a cost function \eqref{eq:cost} with linear-Gaussian priors \eqref{eq:typical-lfm}
for consistency with baseline methods.
We evaluate our method with state-of-the-art algorithms both
for a parametric approach, LatentLasso \cite{Yen2017},
and a Bayesian nonparametric approach with non-negative constraints,
MEIBP \cite{Reed2013}.


We examine our method with both synthetic and actual datasets.
For the synthetic data, we use simulated images also used in \cite{Yen2017}, 
where each feature is $30\!\times\!30$ image and
its randomly selected $7\!\times\!7$ region is set
as\footnote{
  For MEIBP, we use absolute values instead so that non-negative constraint of MEIBP makes sense.
}
$\mathcal{N}(0,1)$.
We employ PDC constraints for the synthetic $Z$ to examine effectiveness of our method for non-identifiability.
For actual-data experiments, we use the {\it UK-DALE}
\cite{UK-DALE} a dataset for Non-intrusive Load Monitoring \cite{Hart1992},
and the {\it Piano} transcription dataset \cite{Piano}.
From the {\it UK-DALE} dataset, we extracted raw current waveforms for every 2 minutes 
from {\it house-1 / 2015 / week-1} data.
And we used {\it Bach\_850} from {\it Piano} dataset
by taking a power spectrum.
Since {\it UK-DALE} data contains negative values and MEIBP is not applicable,
we only applied LatentLasso to it.

Since our method does not change likelihood
except for a small change due to tolerance in Algorithm~\ref{alg:candidate-sampler}.
We evaluate our method by means of following metrics:
\begin{itemize}
  \vspace{-8pt}
  \setlength{\itemindent}{-5pt}
  \setlength{\itemsep}{-3pt}
  \item Hamming Error:
    $ E_\mathrm{Hamm} = \min_{\sigma \in \mathcal{S}_K} \frac{1}{NK}\| Z\sigma - Z^* \|_0$
  \item Regularizer:
    $ E_\mathrm{Reg} = \frac{1}{KD} \| W \|_F$
  \vspace{-6pt}
\end{itemize}
where the first one with ground truth $Z^*$ is only available for synthetic data.
The second corresponds to the logarithm of the prior $P_W$,
which we minimize in Algorithm~\ref{alg:mcmc} via the cost function (\ref{eq:cost})
and it is expected to offer sparser representation of the data,
more representable and closer to true parameters.

\begin{figure}[tb]
  \begin{center}
    \centerline{\includegraphics[width=\textwidth]{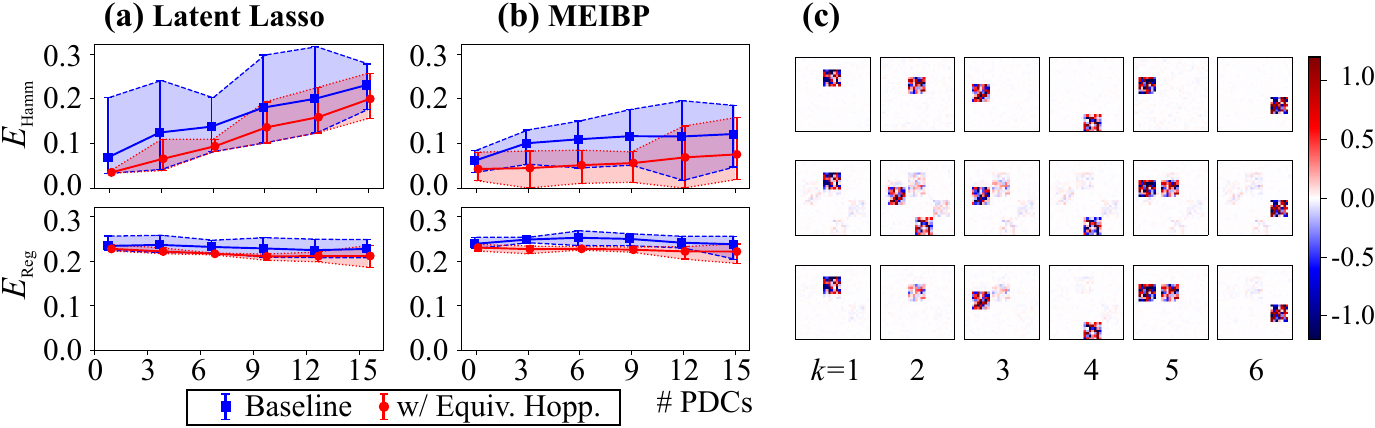}}
    \caption{
      {\bf (a,b)} Metrics for synthetic data ($K=14,N=1000$) with PDC constraints
      estimated by {\bf (a)} LatentLasso and {\bf (b)} MEIBP with/without Equivalence Hopper
      (Algorithm~\ref{alg:mcmc}).
      Markers and error bars represent means and extrema of 5 trials.
      {\bf (c)} Example $W$ for synthetic data ($K=6$) of
      ground truth (upper), estimated by LatentLasso (middle), and with Equivalence Hopper (bottom).
      \label{fig:hamming}
    }
    \vspace{-15pt}
  \end{center}
\end{figure}

\begin{table*}[tb]
\def\mc#1#2#3{\multicolumn{#1}{#2}{#3}}
\begin{center}
\caption{
  Metrics for actual-data experiments.
  The number of features $K^+$ is constant for LatentLasso and estimated variables for MEIBP.
  Residuals $\|X-ZW\|_F$ have not changed in 4 significant digits
  before and after Equivalence Hopper is applied.
  \label{tab:actual}
}
\vspace{-10pt}
\vskip 0.1in
\begin{small}
\begin{sc}
\begin{tabular}{l|lc|cc|cc}
  \toprule
                 &              &       & \mc{2}{c|}{Baseline} & w/ Equiv. Hopp. \\
  Dataset        & Algorithm    & $K^+$ & $\|X-ZW\|_{F}$ & $E_\mathrm{Reg}$ &  $E_\mathrm{Reg}$ \\
  \midrule
  UK-DALE        & LatentLasso  &  5    &  0.422  &  0.692    & 0.677 ({\bf -2.3\%})\\
  UK-DALE        & LatentLasso  & 10    &  0.247  &  1.574    & 1.480 ({\bf -6.0\%})\\
  \midrule
  Piano          & LatentLasso  & 20    &  0.830  &  0.699    & 0.698 ({\bf -0.2\%})\\
  Piano          & LatentLasso  & 40    &  0.698  &  0.492    & 0.491 ({\bf -0.3\%})\\
  Piano          & MEIBP        & 49    &  0.700  &  0.635    & 0.625 ({\bf -1.5\%})\\
  \bottomrule
\end{tabular}
\end{sc}
\end{small}
\end{center}
\vspace{-15pt}
\end{table*}

Figure \ref{fig:hamming} (a,b) shows the metrics for synthetic data with varying number of PDCs ($N_\mathrm{pdc}$).
Accuracies in Hamming error of both two baseline methods (LatentLasso and MEIBP) degrade by increasing $N_\mathrm{pdc}$.
By applying our method, Equivalence Hopper, accuracies are significantly improved,
and especially in MEIBP, the error almost halves in average
and the ground truth have been achieved within 5 trials at $N_\mathrm{pdc} = 3, 12$.
In LatentLasso, while the best-case $E_\mathrm{Hamm}$ is mostly unchanged before and after
applying Equivalence Hopper, a range of $E_\mathrm{Hamm}$ values significantly narrows
keeping the average error lower and the estimation robust.
A robustness of the estimation is quite important for unsupervised learning especially in non-identifiable case,
in which we cannot evaluate the error without knowledge about ground truth.
Further, Equivalence Hopper is worth applying in most cases
rather than repeating the preceding methods until getting better results
since our algorithm never worsen the result,
and executes very fast, delivering results in less than 5 seconds
(less than a single iteration of the preceding methods)
for $M=1000$ iterations with $N_s=1000$.
It is also remarkable that only few percentage improvement of $E_\mathrm{Reg}$
result in a drastic improvement in $E_\mathrm{Hamm}$,
which is consistent with the result in Figure \ref{fig:feasible_space}(b,c)
where
completely different but equivalent solution has small difference in 
$\|W\|_F$.

Example features obtained by LatentLasso with/without Equivalence Hopper are shown in Figure~\ref{fig:hamming} (c).
In a raw estimation of LatentLasso (middle), there are some features, e.g. the second feature,
having multiple features of the ground truth (top) in duplicate.
With application of the Equivalence Hopper algorithm (bottom),
such a duplication was suppressed, and we obtained a sparser solution.
The fourth feature in Fig.~\ref{fig:hamming}(c) is an instance of ``inverted'' feature we discussed in Sec.~\ref{sec:identifiability},
in which the sign of feature is flipped in the result of LatentLasso and is restored by Equivalence Hopper.

Finally, we show the actual-data experiments in Table~\ref{tab:actual}.
Our method obtained a better solution in a manner of a $E_\mathrm{Reg}$
than the preceding algorithm without any degradation
in residual (i.e., likelihood) in all cases.
While the change in $E_\mathrm{Reg}$ is in few percentage,
it cannot be neglected because small change of $E_\mathrm{Reg}$
in synthetic data results in drastic improvement of Hamming error $E_\mathrm{Hamm}$.
We believe that small changes in $E_\mathrm{Reg}$ is important
for comparison between equivalent solutions.


\bibliography{NIPS2018}
\bibliographystyle{IEEEtran}

\newpage

\appendix

\section*{Supplemental Materials}







\section{Remarks on condition $\mathrm{rank}(W)=K$}

While a condition $\mathrm{rank}(Z)\!=\!K$ is
necessary for identifiability as shown in the proof of Theorem~\ref{theo:ident},
a condition $\mathrm{rank}(W)\!=\!K$  is not necessarily required for identifiability.
We can see this by assuming the following counterexample with $K=2, D=1, N=4$:
\begin{equation*}
  Z = \begin{pmatrix}
    0 & 0 \\
    0 & 1 \\
    1 & 0 \\
    1 & 1
  \end{pmatrix},\;
  W = \begin{pmatrix}
    1\\ 10
  \end{pmatrix},\;
  X = ZW = \begin{pmatrix}
    0 \\
    1 \\
    10 \\
    11
  \end{pmatrix},
\end{equation*}
then $(Z,W)$ is identifiable despite $\mathrm{rank}(W) = 1 < K$.

\section{Lower bound for the number of equivalent solutions in biased LFM }
\label{sec:lower-bound}

We show the number of equivalent solutions $|H(Z)/\mathcal{S}_K|$
has a lower bound $(K+1) \, 2^{K-1}$ if $Z$ has a column that is always active.

Assume $k=K$ be a {\it bias} feature (e.g., $\bm{z}_K=\bm{1}$),
and $f$ be as defined in Sec.~\ref{sec:method}.
Then, $f(Z\bm{u}) = 0$ holds for every 
$\bm{u} \in \mathcal{U}
=\{ \bm{e}_1, \cdots, \bm{e}_K, \; \bm{e}_K\!-\!\bm{e}_1, \cdots, \bm{e}_K\!-\!\bm{e}_{K-1}$\},
because $\bm{z}_i,\, \bm{z}_K-\bm{z}_i \in \{0,1\}^{N}$.
Hence,
we can construct $U$ by selecting its columns from $\mathcal{U}$ such that $U$ is regular matrix.

From the regularity of $U$, it includes at least one non-zero element in each row.
Therefore, let
\begin{align*}
  \mathcal{U}_i &= \{ \bm{e}_i, \bm{e}_K - \bm{e}_i \} \;\;\; \mathrm{for} \; i=1,\cdots,K\!-\!1, \\
  \mathcal{U}_K &= \{ \bm{e}_K, \bm{e}_K - \bm{e}_1, \cdots, \bm{e}_K-\bm{e}_{K-1} \},
\end{align*}
then, $U$ is regular iff all $\mathcal{U}_1,\cdots,\mathcal{U}_K$ include
at least one column of $U$.
We count up the number of $U$ by considering two cases:
\begin{itemize}
  \item The case of $U$ including $\bm{e}_K$ in its columns, \\
    the other $K-1$ columns of $U$ are selected one from every $\mathcal{U}_{1}, \cdots, \mathcal{U}_{K-1}$.
    Then, the number of combination is $2^{K-1}$.

  \item The case of $U$ not including $\bm{e}_K$ in its columns,\\
    all $K$ columns of $U$ are selected at least one from every $\mathcal{U}_1, \cdots, \mathcal{U}_{K-1}$.
    Then, two columns of $U$ are selected from a single $\mathcal{U}_k$, and
    the other $K-2$ columns are selected one from every $\mathcal{U}_k' \; (k' \neq k)$.
    Then, the number of combination is $(K-1) \, 2^{K-2}$.
\end{itemize}
Summing up the cases,
we get a lower bound of equivalent solutions:
\begin{equation*}
|H(Z)/\mathcal{S}_K| \geq 2^{K-1} + (K-1) \, 2^{K-2} = (K+1)\,2^{K-2}. \square
\end{equation*}

\section{Sufficient condition for $H(Z) = \tilde{H}(Z)$ }
\label{appendix:HZ_integer}

We show below some sufficient conditions for $\tilde{H}(Z) = H(Z)$.
\begin{theo}
  \label{theo:integer-condition}
  Let $Z \in \{0,1\}^{N \times K}$ be a binary matrix of rank $K$.
  Assume $U \in H(Z)$.
  $U$ is an interger matrix if one of following conditions holds:
  \begin{itemize}
    \item [{\bf a.}] $\min \{ n | z_{n,k} = 1 \} \neq \min \{ n | z_{n,k'} = 1 \}$ for $k\neq k'$,
    \item [{\bf b.}] $Z$ has a $K\times K$ submatrix $\zeta$ s.t. $|\det({\zeta})| = 1$.
  \end{itemize}
\end{theo}

The first condition in Theorem~\ref{theo:integer-condition} assumes that
the timing of the first appearance of each feature
is different from that of each of the others.
In other words, at most one new feature appears at the same time.
The second condition is a more permissive condition
since the first one follows it.
The condition may hold when $N \gg K$
since it holds if at least one of $\binom{N}{K}$ submatrices
in $Z$ has an absolute determinant of one.


\section{Detail of Algorithm \ref{alg:mcmc} for linear Gaussian LFMs}
\label{appendix:mcmc}

In linear-Gaussian LFMs, the cost function \eqref{eq:cost} is represented as:
\begin{equation}
  \label{eq:typical-cost}
  g(U; \hat{Z}, \hat{W}) = \tau \|U^{-1} \hat{W}\|_F^2 + \gamma f(\hat{Z}U),
\end{equation}
We use a MCMC-based method that
samples $U$ according to \eqref{eq:prob}
by updating each column in each iteration.
Using Algorithm~\ref{alg:candidate-sampler}, we sample candidates
$\mathcal{U} = \{\bm{u}^{(1)},\cdots,\bm{u}^{(N_s)} \}$
of the $k$-th column of the next $U'$,
and we select one according to distribution (\ref{eq:prob}) such that $U'$ is regular.
Our method is similar to the multi-try method introduced in \cite{Liu2000, Storvik2011},
but in our method, the proposal of candidates $\mathcal{U}$ does not depend
on the current state of $U$.
This means that we can use the same $\mathcal{U}$
both for sampling proposals and for calculating acceptance ratios,
and this results in the acceptance ratio always being one
if the current $\bm{u}_k$ is among the candidates.
While we could further reuse $\mathcal{U}$ over iterations,
it is better to resample in several iterations for global convergence.

The most time-consuming step in Algorithm~\ref{alg:mcmc} is
singularity determination of
$U' = U + (\bm{u}^{(i)} - \bm{u}_k)\bm{e}_k^\mathrm{T} = U + \Delta \bm{u}^{(i)}\bm{e}_k^\mathrm{T}$
and probability calculation $P(U'|\hat{Z}, \hat{W})$
in \eqref{eq:prob} for every $\bm{u}^{(i)} \in \mathcal{U}$.
For singularity determination, we adapt the {\it rank-1 update} formula of determinants,
\begin{equation}
  \label{eq:detU}
  \det(U') = \det({U})\cdot \left(1 + \bm{e}_k^\mathrm{T} U^{-1} \Delta \bm{u}^{(i)}  \right).
\end{equation}
We can check the singularity of $U'$ by
$
(1 + \bm{e}_k^\mathrm{T} U^{-1} \Delta \bm{u}^{(i)})  $
to be zero.
And for the probability calculation, we get
\begin{equation}
  \label{eq:delta-cost}
  \| U'^{-1} \hat W \|_F^2 = \| U^{-1}\hat  W \|_F^2 
       + \bm{e}_k^\mathrm{T} \, \Omega  \, \left(\|\bm{v}^{(i)}\|^2 \bm{e}_k - 2 \bm{v}^{(i)} \right),
       \nonumber
\end{equation}
where
$
\bm{v}^{(i)} = \frac{ U^{-1} \Delta \bm{u}^{(i)}}{1 + \bm{e}_k^\mathrm{T} U^{-1} \Delta \bm{u}^{(i)}}
$,
and
$\Omega = U^{-1} \hat{W} \hat{W}^\mathrm{T}U^{-\mathrm{T}}$.
Then we finally get
\begin{align}
  \label{eq:delta-prob}
  P&(U'|\hat{Z},\hat{W}) \propto \exp \left\{ -\beta \left(
    \tau \bm{e}_k^\mathrm{T} \Omega \left(\|\bm{v}^{(i)}\|^2 \bm{e}_k - 2 \bm{v}^{(i)}\right)
    +\gamma f(\hat{Z}\bm{u}^{(i)})
  \right) \right\}.
\end{align}
By reusing values of $ f(Z\bm{u}^{(i)}) $ calculated in Algorithm~\ref{alg:candidate-sampler},
the calculation time of (\ref{eq:detU}) and (\ref{eq:delta-prob})
is $\mathcal{O}(K)$ for each candidate if $U^{-1}$ and $\Omega$ are given.
We can maintain these matrices incrementally in $\mathcal{O}(K^2)$ per iteration as 
$U'^{-1} = V U^{-1} $ and $\Omega'^{-1} = V \Omega V^\mathrm{T}$,
where $V=I-\bm{v}^{(i)} \bm{e}_k^\mathrm{T}$.

In total, calculation time with our method is 
$\mathcal{O}(NK^2+DK^2)$
for initializing $\Omega$ and $\Lambda$,
$\mathcal{O}(N_s K^2)$ for resampling candidates (Algorithm~\ref{alg:candidate-sampler}),
and $\mathcal{O}(N_s K + K^2)$ for each MCMC iteration (Algorithm~\ref{alg:mcmc}).
Our method is quite fast since there is no need to scan all $N$ data once initialized.
The convergence of Algorithm~\ref{alg:mcmc} is
proven by following theorem,
\begin{theo}
  \label{theo:mcmc}
  Let $q(U'|U)$ be MCMC kernel defined as Algorithm~\ref{alg:mcmc}. Then,
  \begin{itemize}
    \setlength{\itemindent}{-5pt}
    \setlength{\itemsep}{-2pt}
    \item $q(U'|U)$ satisfies detailed balance condition.
    \item $q(U'|U)$ is transitable over regular integer matrices.
  \end{itemize}
\end{theo}
After $m$ iterations of the MCMC step, the distribution of $U$ will
converge to (\ref{eq:prob}) in $m \to \infty$.

\section{Lower-bound of optimality}

Algorithm 2 has a lower bound on its optimality for typical linear-Gaussian LFMs in a limit of $\beta \to \infty$
because it becomes greedy local search algorithm for submodular maximization on a matroid constraint.
Rewriting the cost function $g$ in \eqref{eq:typical-cost} as:
\begin{equation*}
  g(U; \hat{Z}, \hat{W}) = 
  \tau \mathrm{Tr}[(UU^{T})^{-1} \hat{W}\hat{W}^T] + \gamma \sum_{k=1}^{K} f(\hat{Z} \bm{u}_k),
\end{equation*}
its domain can be extended from regular matrices to rank-K ones with more than K columns.
Assuming $S \subset \mathcal{U}$ be a set of $\bm{u}$ NOT included in columns of U,
then possible $S$'s form a matroid and the target function $g$ is supermodular on it.
So, there exists linear set function $h$ so that $g-h$ is monotone submodular,
and the problem becomes a monotone submodular maximization on a matroid constraint.
It is well known that local search yields 1/2-optimal solution for the problem \cite{Fisher1978}.

\section{Survey on non-identifiability conditions}

We can find out conditions for non-identifiability appears many real-world datasets.
Table \ref{tab:survey} shows the survey on datasets for multi-label classification datasets in LIBSVM library\footnote{
  \url{https://www.csie.ntu.edu.tw/~cjlin/libsvmtools/datasets/multilabel.html}
}.
We can see that at least 6.1\% (siam-competition2007) and
in average 57.3\% of $\frac{1}{2} K(K-1)$ pairs of features
satisfy PDCs in Theorem \ref{theo:pdc}.
These results imply that LFMs will suffer from non-identifiability
in most cases of actual applications.

\begin{table}[tbh]
\caption{ Statistics of datasets in LIBSVM.  }
\label{tab:survey}
\begin{center}
\begin{small}
\begin{sc}
\begin{tabular}{l|rrrr}
  \toprule
  Dataset                 & $N$   & $K$ & \# Pairs of PDC & PDC Ratio \\
  \midrule                              
    mediamill (exp1)      & 30,993 & 101 &  3,074 & 60.9 \% \\
    rcv1v2 (topics)       & 23,149 & 101 &  3,359 & 66.5 \% \\
    rcv1v2 (industries)   & 23,149 & 313 & 46,189 & 94.6 \% \\
    rcv1v2 (regions)      & 23,149 & 228 & 24,925 & 96.3 \% \\
    scene-classification  &  1,211 & 6   &      8 & 53.3 \% \\
    siam-competition2007  & 21,519 & 22  &     14 & 6.1 \% \\
    yeast                 &  1,500 & 14  &      8 & 8.8 \% \\
  \bottomrule
\end{tabular}
\end{sc}
\end{small}
\end{center}
\end{table}

\section{Proofs for theorems}

\subsection*{Theorem \ref{theo:laurberg}}
Already shown in \cite{Laurberg2008}.

\subsection*{Theorem \ref{theo:ident}}

In the case of $\mathrm{rank}(ZW) = K$ (i.e. $\mathrm{rank}(Z)\!=\!\mathrm{rank}(W)\!=\!K$) 
the assertion in Theorem \ref{theo:ident} follows from Theorem \ref{theo:laurberg} as discussed in the paper.
Then we prove in the case of $\mathrm{rank}(Z)\! < \! \mathrm{rank}(W)\!=\!K$,
where both the trivialness and the identifiability is false.
\begin{itemize}
\item Non-trivialness of $H(Z) / \mathcal{S}_K$

  In the case of $Z$ having a column $\bm{z}_l$ which is all zero,
  a matrix $U_0 = I + a\bm{e}_l \bm{e}_l^\mathrm{T}$
  for any $a \neq -1,0$ is in $H(Z)$
  because $\det U_0 = a+1 \neq 0$ and $ZU_0 = Z \in \{0,1\}^{N\times K}$.
  And $U_0$ is not a permutation matrix. Then $|H(Z)/\mathcal{S}_K| \geq 2$.

  In the case that no column in $Z$ is zero vector,
  since $\mathrm{rank}(Z)\! < K$,
  there exists $\bm{b} \in \mathbb{R}^K \backslash \{\bm{0}\}$
  s.t. $Z \bm{b} = \bm{0}$.
  Let $b_m$ be a non-zero component of $\bm{b}$,
  then $\bm{b} \neq \bm{e}_m$
  because $Z \bm{b} = \bm{0}$ and $Z \bm{e}_m =  \bm{z}_m \neq \bm{0}$.
  A matrix $U_1 = (\bm{e}_1, \cdots, \bm{e}_{m-1}, \bm{b}, \bm{e}_{m+1}, \cdots, \bm{e}_K)$
  is in $H(Z)$ because
  $\det U_1 = b_m \neq 0$ and
  $ZU_1 = (\bm{z}_1,\cdots,\bm{z}_{m-1},\bm{0},\bm{z}_{m+1},\cdots,\bm{z}_K) \in \{0,1\}^{N\times K}$.
  And $U_1$ is not a permutation matrix. Then $|H(Z)/\mathcal{S}_K| \geq 2$.
  
\item Non-identifiability of $(Z,W)$

  In the case of $Z$ having a column $\bm{z}_l$ which is all zero,
  $(ZU_0, U_0^{-1}W)$ is another solution 
  since $U_0^{-1}W \neq W$ (whereas $ZU_0 = Z$).

  In the case that no column in $Z$ is zero vector,
  $(ZU_1, U_1^{-1}W)$ is another solution 
  because $ZU_1 = (\bm{z}_1,\cdots,\bm{z}_{m-1},\bm{0},\bm{z}_{m+1},\cdots,\bm{z}_K) \neq Z$.
  $\square$

\end{itemize}

\subsection*{Theorem \ref{theo:iid}}

Assume $\{ Z_{n,:} \}_{n=1}^{N} = \{0,1\}^{K}$ and $U \in H(Z)$.
Let $\bm{u}_k^T$ be a $k$-th row or $U$.
Since $\bm{e}_1^\mathrm{T}, \cdots, \bm{e}_K^\mathrm{T}$ and $\bm{1}^\mathrm{T}$ is
included in rows of $Z$, i.e.
$$
{}^\exists n_1, \cdots, n_K, m, \;\;\; Z_{n_k,:} = \bm{e}_k, Z_{m,:} = \bm{1},
$$
then
$$
\bm{u}_k = (\bm{e}_k^\mathrm{T}U)^\mathrm{T}  = (Z_{n_k,:}U)^\mathrm{T} \in \{0,1\}^{K},
$$
$$
\sum_{k=1}^K \bm{u}_k = (\bm{1}^\mathrm{T}U)^\mathrm{T} = (Z_{m,:}U)^\mathrm{T} \in \{0,1\}^{K}.
$$
Therefore, $U$ is binary matrix and each row of $U$ sums to one.
Also consider that $U$ is regular, $U$ is permutation matrix.
Hence $H(Z) = \mathcal{S}_K$. $\square$



\subsection*{Theorem \ref{theo:pdc}}

We provide examples of transfer matrices.
Let $\bm{e}_i$ be a vector of which the $i$-th component is 1 and the others are 0, and
\begin{equation*}
  \label{eq:QR}
  R_{ij} = I + \bm{e}_i \bm{e}_j^\mathrm{T}, \;
  Q_{ij} = I + \bm{e}_i \bm{e}_j^\mathrm{T} - 2 \bm{e}_i \bm{e}_i^\mathrm{T}.
\end{equation*}
Then 
$I, R_{ij}, R_{ji} \in H(Z)$ in PDC1,\,
$I, R_{ij}^{-1}, Q_{ji} \in H(Z)$ in PDC2,\, and  \,
$I, Q_{ij}, R_{ji}^{-1} \in H(Z)$ in PDC3. $\square$

Note that PDC2 and PDC3 are essentially the same because one can be derived from the other
by employing contraposition and exchanging $i$ and $j$.

\subsection*{Theorem \ref{theo:integer-condition}}

We firstly prove the case of condition {\bf b.},
and then prove {\bf a.} by using it.

\begin{itemize}
  \item[{\bf b.}] $Z$ has a $K\times K$ submatrix $\zeta$ s.t. $|\det({\zeta})| = 1$.
    
    Assume $U \in H(Z)$.
    Let $\zeta'$ be a $K \times K$ submatrix of $Z'=ZU$
    picking the same rows as $\zeta$.
    Then $\zeta' = \zeta U$.

    As $\zeta$ is a binary matrix (hence, an integer matrix) and $|\det({\zeta})| = 1$,
    $\zeta$ is a unimodular matrix,
    and it has an inverse $\zeta^{-1}$ that is also a unimodular matrix.
    Hence $U = \zeta^{-1}\zeta'$ is integer matrix.



  \item[{\bf a.}] $\min \{ n | z_{n,k} = 1 \} \neq \min \{ n | z_{n,k'} = 1 \}$ for $k\neq k'$.

    Let $m(k) = \min \{ n | z_{n,k} = 1 \}$, and $m(k)$ is well-defined because
    $\{ n | z_{n,k} = 1 \}$ is not empty since $\mathrm{rank}\,Z=K$.
    The condition assumed states that $m(k)$ is injective.

    Consider a integer array $(l_1,\cdots,l_K)$
    that is sorted from $(1,\cdots,K)$ by ascending order of $m(l_k)$ values.
    Let $K \times K$ submatrix $\zeta$ of $Z$ as $ \zeta_{k,:} = Z_{m(l_k),:}$
    (i.e. $k$-th row of $\zeta$ is a row of $Z$ where the $k$-th earliest appeared feature appears).
    Then $\zeta$ is a lower triangular matrix with diagonal components of one, having
    $\det \zeta = 1$.
    Hence $U\in \mathbb{Z}^{K\times K}$ follows from {\it c.} $\square$

\end{itemize}

\subsection*{Theorem \ref{theo:mcmc}}

\subsubsection*{Detailed Balance Condition}

To prove the convergence of Algorithm~\ref{alg:mcmc},
we first prove the detailed balance condition, a sufficient condition
for MCMC kernel $q(U'|U)$ to keep the intended distribution $P(U)$ invariant.

We factorize the MCMC kernel $q(U'|U)$ as
$$
q(U'|U) = q(U'|U, k, \mathcal{U}) \, q(k) \, q(\mathcal{U})
$$
where $q(\mathcal{U})$ is a probability to obtain $\mathcal{U}$ by Algorithm~\ref{alg:candidate-sampler},
$q(k) = 1/K$, and
$$
q(U'|k, U, \mathcal{U}) =
\frac
{ \mathbb{I} \left[U'\in D_k(U,\mathcal{U}) \right] \cdot P(U') }
{ \sum_{U' \in D_k(U, \mathcal{U})} P(U') } ,
$$
where
\begin{align*}
  D_k&(U, \mathcal{U}) \\\
     &= \left\{ U + \Delta \bm{u} \middle|
        \bm{u}_k + \Delta \bm{u} \in \mathcal{U},
        \det ( U + \Delta \bm{u}) \neq 0 \right\}.
\end{align*}

We assume $x = (k, \mathcal{U})$ as {\it auxiliary variables}, and use
Proposition~4 in \cite{Storvik2011} with
$$
h(x| U, x^*, U') = \delta(k,k^*) \, \delta(\mathcal{U}, \mathcal{U}^*),
$$
and get an acceptance ratio
\begin{align*}
  r(U;x^*,U',x)
  &= \frac{P(U') \, q(U|U', k^*, \mathcal{U}^*)}
          {q(U'|U, k, \mathcal{U}) \, P(U)}  \\
  &= \begin{cases}
        1, &  \bm{u}_k \in \mathcal{U}, \\
        0, &  otherwise.
     \end{cases}
\end{align*}
Then Algorithm~\ref{alg:mcmc} satisfies a detailed balance condition. $\square$

\subsubsection*{Transitability}

Since a value of $q(U'|U)$ becomes zero for some $U'$, we need to prove transitability
for global convergence.
By transitability we meant that:
for arbitrary $U$ and $U^{(0)}$,
a probability $Q_T(U|U^{(0)})$ to obtain $U$
from initial state $U^{(0)}$ after some finite $T$ MCMC iteration is non-zero.

Assume $U$ and $U^{(0)}$ be integer regular matrices.
Let $V, V_0$ be a set of columns of $U, U^{(0)}$, respectively.
If $ V \neq V_0$, we select $\bm{v}_0 \in V_0$ that is not in $V$.
There exists $\bm{v}_0' \in V$ such that $\dim \mathrm{span}(V_0 \backslash \{\bm{v}_0\} \cup \{\bm{v}_0'\}) = K$,
because if there is no, ${}^\forall \bm{v}' \in V, \bm{v}' \in \mathrm{span}(V_0 \backslash \{\bm{v}_0\})$
yields $\dim \mathrm{span} V = \dim \mathrm{span}(V_0 \backslash \{\bm{v}_0\}) = K-1$
and it conflicts with $\dim \mathrm{span} V = \mathrm{rank}(U) = K$.
Then we set $V_1=V_0\backslash\{\bm{v}_0\}\cup \{\bm{v}_0'\}$.
Repeating above operation while $V_i \neq V$,
we obtain a sequence
$V_0, V_1, \cdots, V_L=V$,
where $L \leq K$ and $V_i\backslash\{\bm{v}_i\} = V_{i+1}\backslash \{\bm{v}_i'\}$.

If the distribution of $f^*$ in Algorithm~\ref{alg:candidate-sampler}
has a support covering $\{0,1,2,\cdots\}$, a probability of $\mathcal{U}$ to 
include both $\bm{v}_i, \bm{v}'_i$ is non-zero.
Therefore, in the $i$-th MCMC step,
a probability to obtain $U^{(i)}$ (which have column entries $V_i$)
from $U^{(i-1)}$ is non-zero.
Hence after $L$ iteration, the probability $Q_L(U',U^{(0)})$
for $U'$ having the same column entries as $U$ is non-zero.

Finally we consider a pseudo operation shuffling columns of $U^{(i)}$ after each iteration
(which is no effect at all, and even no need to be executed),
we get $Q_L(U, U^{(0)}) > 0$ for all regular integer matrices $U, U^{(0)}$. $\square$

\end{document}